\documentclass[11pt, oneside]{article}   	
\usepackage{geometry}                		
\usepackage{graphicx}				
\usepackage{url,amsmath,amssymb,epsfig,subcaption,mathtools,amsthm}

\theoremstyle{definition}

\theoremstyle{remark}

\title{Doing the impossible: Why neural networks can be trained at all}
\author{Nathan Hodas, Panos Stinis\\
Pacific Northwest National Laboratory, WA 99354}
\date{}							

\begin{document}
\maketitle

\begin{abstract}
As deep neural networks grow in size, from thousands to millions to billions of weights, the performance of those networks becomes limited by our ability to accurately train them. A common naive question arises: if we have a system with billions of degrees of freedom, don't we also need billions of samples to train it?  Of course, the success of deep learning indicates that reliable models can be learned with reasonable amounts of data. Similar questions arise in protein folding, spin glasses and biological neural networks. With effectively infinite potential folding/spin/wiring configurations, how does the system find the precise arrangement that leads to useful and robust results?  Simple sampling of the possible configurations until an optimal one is reached is not a viable option even if one waited for the age of the universe. On the contrary, there appears to be a mechanism in the above phenomena that forces them to achieve configurations that live on a low-dimensional manifold, avoiding the curse of dimensionality. In the current work we use the concept of mutual information between successive layers of a deep neural network to elucidate this mechanism and suggest possible ways of exploiting it to accelerate training. We show that adding structure to the neural network that enforces higher mutual information between layers speeds training and leads to more accurate results.  High mutual information between layers implies that the effective number of free parameters is exponentially smaller than the raw number of tunable weights. 

\end{abstract}

\section{Introduction}\label{}

Artificial neural networks with millions, or even billions~\cite{shazeer2017outrageously}, of weights provide neurons and synapses comparable with computational complexity approaching small animals~\cite{Goodfellow-et-al-2016}.  And, scientists have begun using them to test and compare many hypotheses in cognitive science~\cite{phillips2017assessing}. Some work has begun to explore how these complex systems reach such finely balanced solutions.  For example,  some have addressed how, given that the space of possible functions is so large, can any finite computational stage do a good job approximating physical systems~\cite{lin2017does}. However, from a cognitive science perspective, the converse question remains, how is it that these complex systems can be trained with only a reasonable amount of data (vastly less than the complexity of the systems would suggest)? Given the computational power available in modern GPUs, we may explore these artificial neural networks to better understand how such highly interconnected computational graphs transfer information to quickly reach global optima.

Deep neural networks have shown great promise in a host of machine learning tasks in computer vision, speech recognition and natural language processing (see e.g. the review \cite{lecun2015deep} and references therein). Exactly because of this success, there exists a need to understand what sets deep learning apart from other approaches, explain how it can achieve the impressive results that have been recently documented, identify the limitations  and investigate more efficient designs within the restrictions. 

Deep neural networks have grown in size, from thousands to millions to billions of weights and the performance of those networks becomes limited by our ability to accurately train them \cite{srivastava2015highway, klambauer2017selfnormalizing}.    Thus, a question that arises is: if we have a system with billions of degrees of freedom, don't we also need billions of samples to train it?  The success of deep learning indicates that reliable models can be learned with reasonable amounts of data. Similar behavior appears in protein folding, spin glasses and biological neural networks.  

In the case of protein folding, there is a vast number of conformations that the protein can assume which do not correspond to a folded state. Simple statistical sampling of the configurations would take astronomically long times to find the folded state. Yet, when the protein starts to fold it completes this task relatively fast (see also Levinthal paradox \cite{dill}). The resolution lies in the fact that evolution has created a mechanism of folding which involves the rapid formation of local interactions. These interactions determine the further folding of the protein. The mechanism can be described by a funnel-like energy landscape \cite{dill}. The funnel-like energy landscape has deep, steep walls with intermediate plateaus. This drastic landscape correlates most of the degrees of freedom and allows the protein folding to proceed in relatively few, large steps towards its folded state.  

The training of deep neural networks involves an optimization problem for the parameters (weights) of the network. Recent work \cite{choromanska} has used the fact that the loss function involved in the optimization can be mapped to a spin glass model in order to study the landscape of the loss function. In particular, it was found that the landscape contains a large number of local minima whose number increases exponentially with the size of the network. Most of these local minima are equivalent and thus yield similar performance on a set of test samples. While the existence of a lot of (mostly) equivalent local minima explains the common behavior of deep neural network training observed by different researchers, we want to study in more detail the approach to the minima. It is known that these minima can be highly degenerate which makes the picture of local funnel-like energy landscapes more plausible (see also previous paragraph about protein folding). This local funnel-like energy landscape picture points towards the notion that, during training, the neural network is able to achieve configurations that live on a low-dimensional manifold, avoiding the curse of dimensionality. Thus, we want to study {\it how} the interplay of depth, width and architecture of the network can force it to achieve configurations that live on that manifold.

The restriction to a low-dimensional manifold is facilitated by the contractive properties of popular activation functions or regularization techniques. But this is not enough to explain why the deep neural nets work well and more importantly how to train them efficiently. History has shown that, until very recently, adding depth impeded effective training, regardless of the number of training epochs \cite{srivastava2015highway, klambauer2017selfnormalizing}. We will show that deep nets work exactly because they learn features of the data gradually, i.e., in succession starting from simple to more complicated ones. It is known that convolutional neural nets learn features of higher and higher semantic complexity at each layer, but, more precisely, the net finds the correct low-dimensional manifold on which to build the representation of the desired function of the data. The features of the lower layer constrain the space of possible features in the deeper layers. The realization of the need for gradual learning of features suggests, in mathematical terms, that the successive layers of the deep net should be highly correlated and that highly-nonlinear activation functions that destroy correlation will impede training of large networks. We show how this concept is connected to a number of emerging training techniques, such as batch normalization and ResNets. It is also related to the recently pointed connection between the Variational Renormalization Group and Restricted Boltzmann Machines \cite{mehta} as well as the Information Bottleneck analysis of deep neural networks \cite{tishby2017}. We compare the layer-by-layer feature learning of nets where correlation between layers is enforced and those without it. Lastly, we discuss how these ideas form promising design principles for more efficient training of neural nets of high complexity.

\section{Materials and Methods}\label{materials_methods}

To evaluate the learning process of the neural networks, we created and trained numerous neural networks.  To create the neural networks, we used Keras~\cite{chollet2015keras} with a TensorFlow~\cite{abadi2016tensorflow} back-end.  We selected a well understood, yet non-trivial, machine learning task: MNIST ({\url{http://yann.lecun.com/exdb/mnist/}), which is identifying hand-written digits.    Each image is a 28x28 greyscale image of a hand-written digit between 0-9.  There are 60,000 training examples and 30,000 validation examples. The validation error (denoted test error in the figures) is the proportion of validation examples the network incorrectly labels.  Recent neural networks have been able to accurately identify over 99.5\% of the validation examples correctly~\cite{chang2015batch}.  However, MNIST is non-trivial, as these excellent results were only achieved in recent years using deep learning.  Thus, poor training will produce poor results on MNIST, while good training will provide excellent results.  MNIST has a large enough input space (784 pixels) to present a challenge, but small enough to tractably explore many training configurations with a single GPU. Training was conducted on a NVidia 1080p GPU.

We chose to use multilayer perceptrons (MLPs), with and without residual connections.  MLPs are traditional multilayer neural networks, where information flows from one densely connected layer to the next, lastly passing through a softmax layer to provide the prediction of the input digit.  Residual connections alter the topology of the MLPs by adding skip connections, which add shortcuts between layers~\cite{he2016deep}. Explicitly, a ``skip connection" works as follows.  Consider a receiving layer $R$ in a multi-layer perceptron and two other layers $R_1$ and $R_k$ that have a common width w and are both closer to the input than R. The skip connection idea is implemented as follows: For each unit in $R_1$ and the corresponding unit in $R_k$, their output values are added together to create a single combined activation output value so that the total input to layer $R$ is a vector whose elements are equal to the unit-wise sum of the output of layer $R_1$ and the output of layer $R_k$.

We tested  neural networks of different widths (more neurons per layer) and different depths (more layers).  A wider neural network has more expressive power than a narrow neural network, and a deeper neural network has greater expressive power than a shallow one~\cite{eldan2016power,poole2016exponential}. By comparing different widths and depths, we can compare the effects of successive transformations of the data. 

We also added Batch Normalization between each layer, followed by a $\tanh$ activation function.  Batch Normalization corrects for covariate shift due to diverging activations, and it improves the trainability of the model.  Batch Normalization allows us to construct deep neural networks that don't suffer from vanishing gradients like vanilla neural networks~\cite{ioffe2015batch}.  The reduction in vanishing gradients allows us to focus our analysis on the effects of successive information transformations, and not artifacts due to finite numerical precision and training time. 

More specifically, to implement the shortcut connection, we sum the output of the first layer and penultimate layer before passing the sum into the final softmax layer as described above.   To implement the residual connections, we summed the output of alternating layers of the same width, using a topology illustrated in Fig.~\ref{fig:3}. We initialized the networks by sampling from a  normal distribution modified according to~\cite{pmlr-v9-glorot10a,he2016deep}, which has shown to produce weights that promote faster convergence by preventing gradients from starting out pathologically small.  We chose a categorical (softmax) cross-entropy as the objective function. We trained using RMSprop with an initial learning rate of $10^{-3}$. We chose tanh activation functions, and we used the out-of-the-box ``BatchNormalization" layer implemented by Keras ({\tt https://keras.io/layers/normalization/\#batchnormalization}). We used a batch-size of 60,000, meaning all training images were combined into a single back-propagation step. 


The question about what ``gradual learning" means can be partially addressed through the concept of mutual information. The mutual information between two distributions $X$ and $Y$ is defined as
\begin{align*}
MI(X;Y)=H(X)-H(X|Y)=H(Y)-H(Y|X)
\end{align*}
where $H(X)$ and $H(Y)$ are marginal entropies while $H(X|Y)$ and $H(Y|X)$ are conditional entropies. Mutual information is the amount of uncertainty, in bits, reduced in a distribution X by knowing Y.  It is symmetric, meaning $MI(X;Y) = MI(Y;X)$.  It is also invariant to isomorphic transformations, so $MI(X;Y) = MI(g(X);h(Y)),$ for arbitrary invertible functions $g$ and $h$. These properties make mutual information useful for quantifying the similarity between two nonlinearly different layers. It will capture the information lost by sending information through the network, but, unlike traditional correlation measures, it does not require a purely affine relationship between $X$ and $Y$ to be maximized. We calculate the mutual information between the features of two layers by using the Kraskov method~\cite{kraskov2004estimating} using the NPEET Python library ({\tt https://github.com/gregversteeg/NPEET}).  In particular, we take an input image and evaluate the activations at each layer.  We then calculate the mutual information between the activations of the first layer and the last layer, using the entire validation set as an ensemble.   To ensure that the mutual information between the first and last layer is not trivial, we make the first and last layers twice as wide, to force the network to discard information between the first and last layer. 

\begin{figure}[h!]
\begin{center}
\includegraphics[width=7cm]{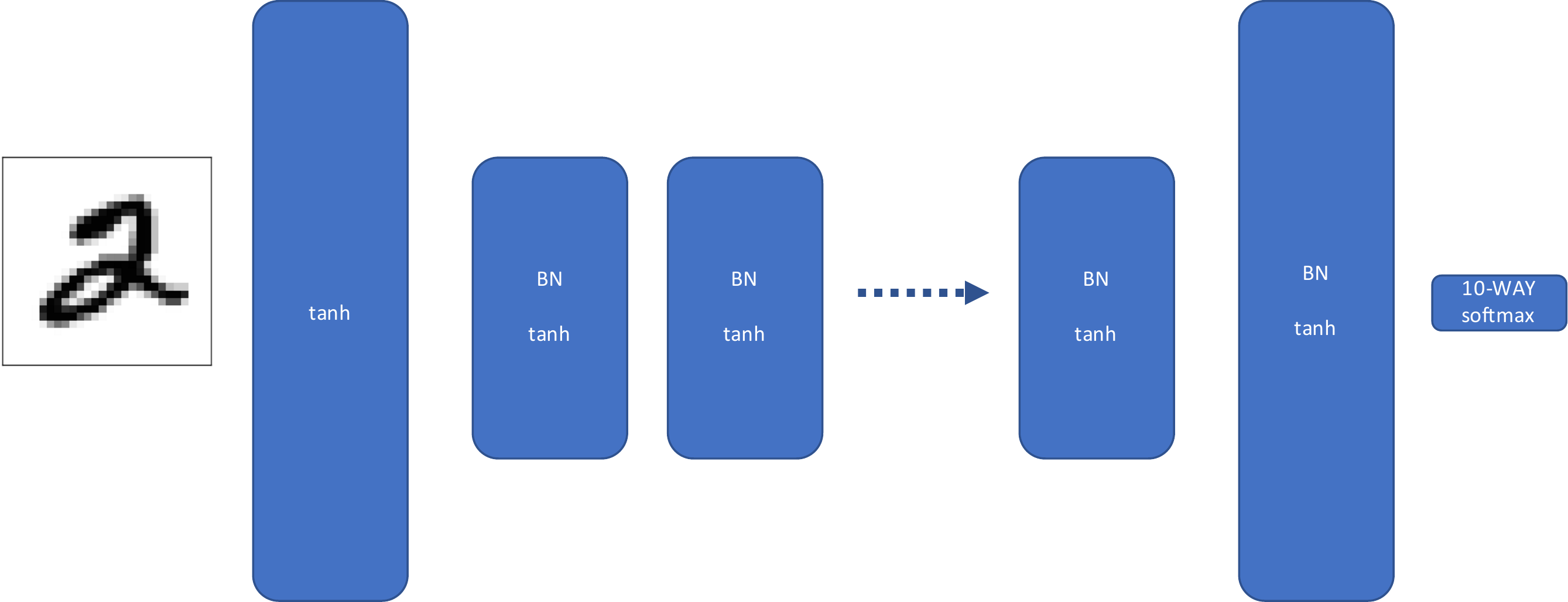}
\end{center}
\caption{ Traditional multilayer perceptron (MLP). Our MLP  consists of multiple layers of width W, where each layer is batch normalized and given a tanh activation.  The first and last layers are twice as wide, to force the network to discard information between the first and last layer.
 }\label{fig:1}
\end{figure}

\begin{figure}[h!]
\begin{center}
\includegraphics[width=7cm]{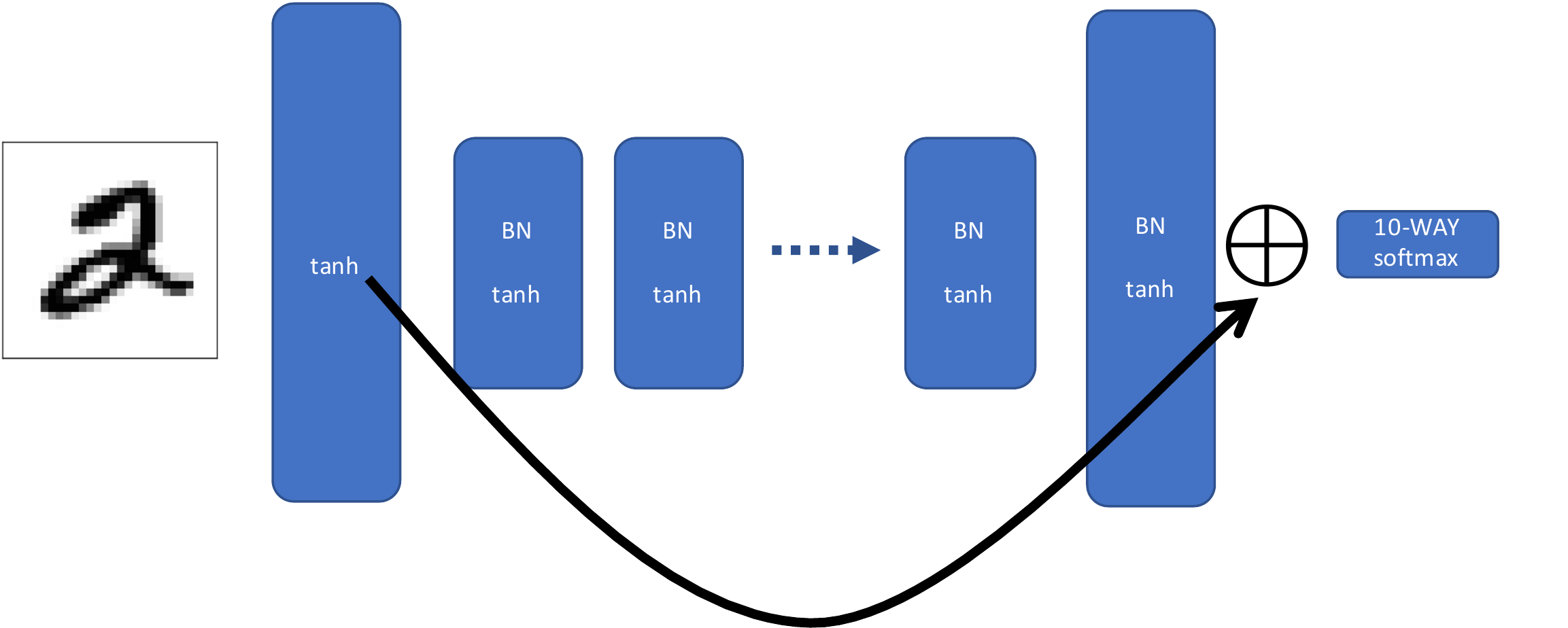}
\end{center}
\caption{ Shortcut network. This is the same MLP structure as in Fig. \ref{fig:1}, but the output of the first and last layer are summed together and fed into the softmax. That is, for each unit in the first and last layer, the two output values are added together to create a single combined activation vector the same size as the (identical) widths of the first and last layers. The shortcut network allows information during backpropagation to propagate the entire length of the network in a single iteration.
 }\label{fig:2}
\end{figure}

\begin{figure}[h!]
\begin{center}
\includegraphics[width=7cm]{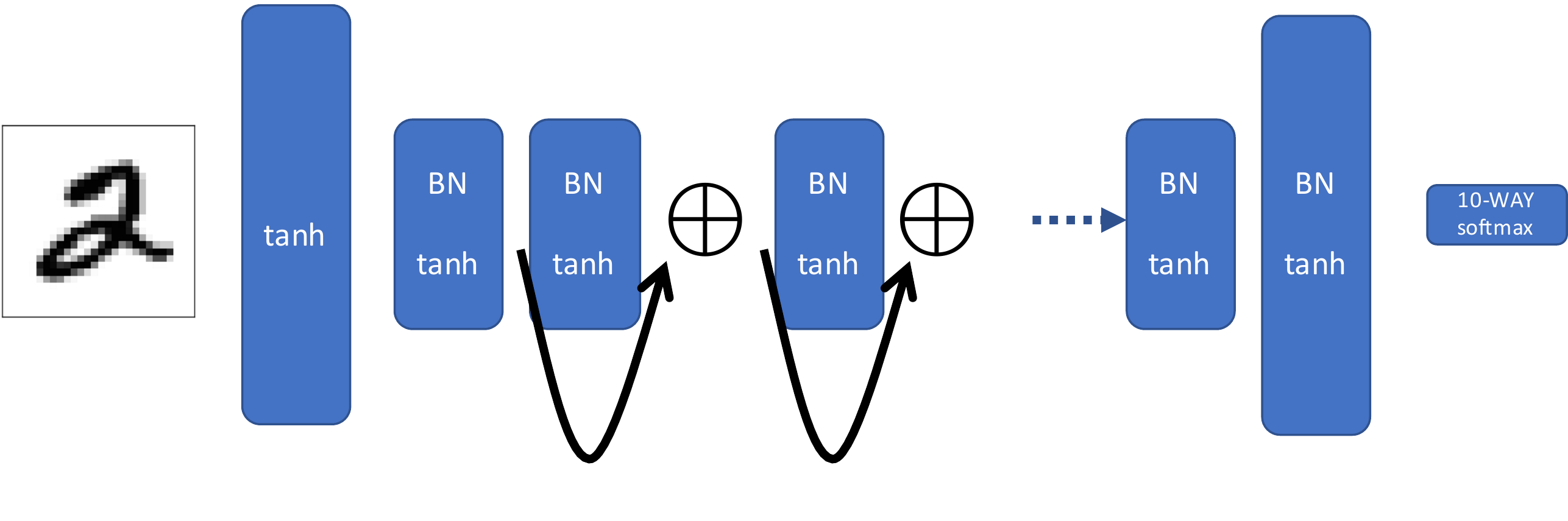}
\end{center}
\caption{ Residual network. The outputs of alternating layers are summed, causing a shortcut between every other layer. Information via backpropagation flows more efficiently backwards into the network, but it can not jump as far in each iteration as the shortcut network.
 }\label{fig:3}
\end{figure}

\section{Results}\label{results}

As shown in Figs. \ref{fig:4a} and \ref{fig:4b}, as the nets train, they progressively move toward an apparent optimum mutual information between the first and last layers.  Traditional MLPs follow a trend of systematically increasing the mutual information. On the other hand, MLPs with shortcuts start with higher mutual information which then decreases towards the optimum. This may be interpreted as the shortcut helping the network to first find a low dimensional manifold, and then progressively exploring larger and larger volumes of state-space without losing accuracy. We should note that the purpose of this study is not to present the state of the art results  (e.g. see \cite{pmlr-v28-wan13} about image classification but the {\it relative} advantage of an architecture using shortcuts over one that does not.

\begin{figure}[htbp]
\begin{minipage}[b]{.5\linewidth}
\centering\includegraphics[width=8cm]{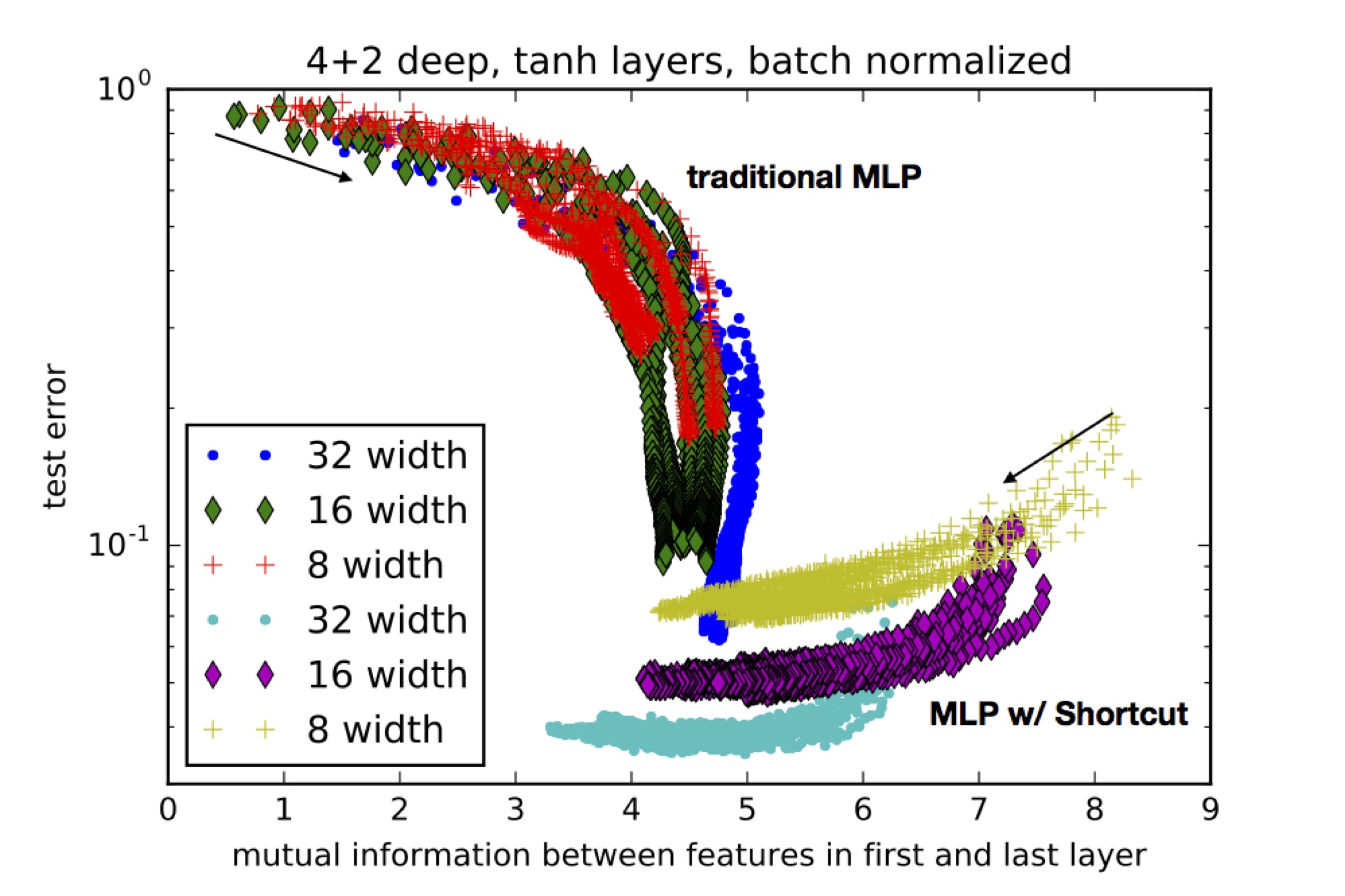}
\subcaption{}\label{fig:4a}
\end{minipage}%
\begin{minipage}[b]{.5\linewidth}
\centering\includegraphics[width=7.5cm]{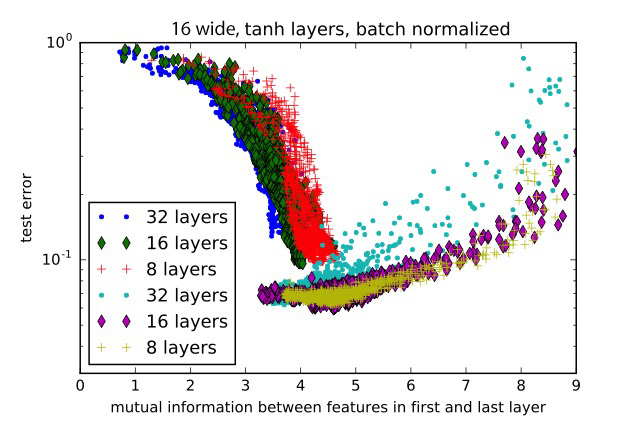}
\subcaption{}\label{fig:4b}
\end{minipage}
\caption{Comparison of performance for nets with a) various layer widths and b) various number of hidden layers.  Each trace represents a different random weight initialization.  The networks with a shortcut are labeled as ``MLP w/ Shortcut.'' The arrows indicate the general direction of progression during training.  Test error is the proportion of validation examples the network incorrectly labels.}\label{fig:4}
\end{figure}

In Figs. \ref{fig:5a} and \ref{fig:5b} we compare the performance of different ResNets widths and the effects of adding residual skip connects, shortcuts, or both respectively. As ResNets train, they start with low mutual information between weights.  The MI gradually increases as it trains, maximizes and begins to decrease again (see Fig. \ref{fig:5a}). The lack of mutual information in the final trained networks shows that a well trained  network does not learn identity transforms.  Also, as can be seen in Fig. \ref{fig:5b}, residuals help both the traditional MLP and shortcut reach a higher final accuracy.   In Fig.~\ref{fig:5a} we see evidence that high mutual information is not a necessary condition for accuracy. However, high mutual information allows the weights to lie upon a low-dimensional manifold that speeds training.  In Fig. \ref{fig:5a}, we see that high mutual information produces rapid decrease in test error: The points that  represent the outcome of each epoch of training show a high slope (and decrease in error) at high mutual information, and a low slope at low mutual information (In \ref{fig:5b}, notice that the x-axis has a different scale). This behavior agrees with the analysis in \cite{tishby2017} which identifies two phases in the training process: i) a {\it drift} phase where the error decreases fast (while the successive layers are highly correlated) and ii) a {\it diffusion} phase where the error decreases slowly (if at all) and the representation becomes more efficient.  The training progress of networks (both MLP and ResNets) with shortcut connections,  indicated by the larger turquoise circles and green crosses, starts with such a high mutual information that the networks are largely trained within a single epoch.

\begin{figure}[htbp]
\begin{minipage}[b]{.5\linewidth}
\centering\includegraphics[width=8cm]{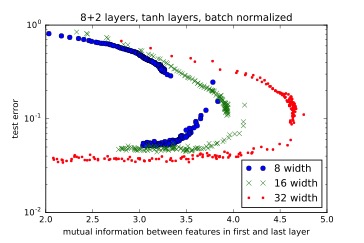}
\subcaption{}\label{fig:5a}
\end{minipage}%
\begin{minipage}[b]{.5\linewidth}
\centering\includegraphics[width=8cm]{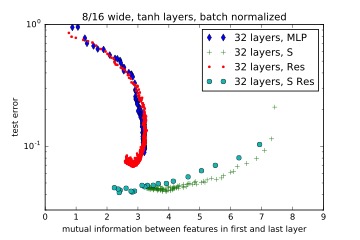}
\subcaption{}\label{fig:5b}
\end{minipage}
\caption{Comparison of performance for a) various ResNet widths and b) various architectures. MLP: Multilayer Perceptron, S: Shortcut, Res: Residual Network, S Res: Shortcut Residual Network. In this plot, as neural networks train, they start at high error and progressively decrease error after each epoch (represented by each point).}\label{fig:5}
\end{figure}

Successive layers which enjoy high mutual information obviously  learn features that cannot be far from the previous layer in the space of possible features. However, mutual information alone cannot tell us \textit{what} these features are. In other words, while we see that the deep net must be learning slowly we cannot use solely mutual information to say what it is that it learns first, second, third etc. This is particularly evident in our observation that training first correlates features in different layers, and then the mutual information steadily decreases as the network fine-tunes to its final accuracy. Thus, we see that high mutual information between layers (particularly between the first and last layer) allows the neural network to quickly find a low dimensional manifold of much smaller effective dimension than the total number of free parameters.  Gradually, the network begins to explore away from that manifold as it fine tunes to its final level of accuracy.

The gathered experience by us and others about the difficulty of training deep nets over shallow nets points to the fact that the first features learned have to be simple ones. If not, if the complicated features were the ones learned through the first few layers, then the deeper layers would not make much difference. But they do, so the ``gradual" learning of features must involve a gradual progression through the space of features from simple to more complex. Another way to think of this is that the depth of the deep net allows one to morph a representation of the input space from a rudimentary one to a sophisticated one. This makes mathematical, physical and evolutionary sense too (see also the analysis in \cite{tishby2017}). This point of view agrees with the success of the recently proposed ResNets. ResNets enforce the gradual learning of features by strongly coupling successive layers. This approach agrees also with the recent realization that Restricted Boltzmann Machines have an exact mapping to the Variational Renormalization Group (vRNG) \cite{mehta}. In particular, in vRNG one proceeds to estimate the conditional probability distribution of one layer conditioned on the previous one. This task is made simpler {\it if} the two successive layers are closely related. In machine learning parlance, this means that the two successive layers are coupled so that the features learned by one layer do not differ a lot from those learned by the previous one. This also chimes with the recent mathematical analysis about deep convolutional networks \cite{mallat}.

The question about what are the ``best" deep net architectures (for a fixed number of layers) can be also partially addressed through measuring mutual information. In particular, tracking the evolution of mutual information and the associated test error with the number of iterations helps us delineate which architectures will find the optimal mutual information manifold, something one should keep in mind when fiddling with the myriads of possible architecture variants. However, mutual information alone is not enough, because it can help evaluate a given architecture but cannot propose (suggest) a new architecture. An adaptive scheme which can create hybrids between different architectures is some kind of remedy but of course does not solve the problem in its generality. This is a well-known problem in artificial intelligence and for some cases it may be addressed through techniques like reinforcement learning \cite{sutton}.

Overall, the successful training of a deep net points to the successful discovery of a low-dimensional manifold in the huge space of features and using it as a starting point for further excursions in the space of features. Also, this low-dimensional manifold in the space of features constrains the weights to also lie in a low-dimensional manifold. In this way, one avoids being lost in unrewarding areas and thus leads to robust training of the deep net.  Introducing long-range correlations appears to be an effective way to enable training of extremely large neural networks.  Interestingly, it seems that maximizing mutual information does not directly produce maximum accuracy, but finding a high-MI manifold and from there evolving towards a low-MI manifold allows training to unfold more efficiently.

\section{Discussion}\label{discussion}

When the output of two layers is highly correlated, many of the potential degrees of freedom collapse into a lower dimensional manifold due to the redundancy between features.  Although we say 'correlation,' we precisely measured this redundancy using mutual information, which is invariant under arbitrary invertible nonlinearities. High mutual information implies that the effective size of the available training state-space has been reduced on order $\sim2^{n*MI}$, where $n$ is the number of layers.  Thus, high mutual information between the first and last layer enables effective training of deep nets by exponentially reducing the size of the potential training state-space. 

Despite having millions of free parameters, deep neural networks can be effectively trained.  How?  We showed that significant inter-layer correlation (mutual information) reduces the effective state-space size, making it feasible to train such nets.  By encouraging the correlation with shortcuts, we reduce the effective size of the training space, and we speed training and increase accuracy. Hence, we observe that long range correlation effectively pulls systems onto a low-dimensional manifold, greatly increasing tractability of the training process.  Once the system has found this low-dimensional manifold, it then tends to gradually leave the manifold as it finds better training configurations.  Thus, high correlation followed by de-correlation appears to be a promising method for finding optimal configurations of high-dimensional systems. By experimenting with artificial neural networks, we can begin to gain insight into the developmental processes of biological neural networks, as well as protein folding \cite{dill}.

Even when batch normalization is used to help eliminate vanishing gradients, deep MLPs remain difficult to train.  As we see in Figure \ref{fig:4b}, beyond 5 to 10 layers, adding depth to a MLP slows training and converges to a lower accuracy.  This has also been demonstrated in other applications with other types of neural networks~\cite{srivastava2015highway}.  Our measures of mutual information also show that deeper networks reduce mutual information between the first and last layer, increasing the difficulty for the training to find a low-dimensional manifold to begin fine tuning. The present results imply that the power of residual networks lies in their ability to efficiently correlate features via backpropagation, not simply in their ability to easily learn identity transforms or unit Jacobians.

The shortcut architecture we describe here is easy to implement using  deep learning software tools, such as Keras or TensorFlow. Despite adding no new free parameters, the shortcut conditions the network's gradients in a way that accelerates training and ultimately increases accuracy.  This follows from the nature of the backpropagation algorithm:  error in the final output of the neural network is translated into weight updates via the derivative chain rule. Adding a shortcut connection causes the gradients in the first layer and final layer to be  summed together, forcing their updates to be highly correlated.  Adding the skip connection increases coupling between the first and final layer, which constrains the variation of weights in the intervening layers, driving the space of possible weight configurations onto a lower dimensional manifold. Thus, a contribution of understanding that the neural networks train more effectively when they start on a low dimensional manifold includes demonstrating how long range shortcuts improve network trainability.  As networks grow in complexity, adding shortcut connections will help keep them on a low dimensional manifold and accelerate training and potentially increase accuracy.

Thus, toward the central question of 'how can neural networks be trained with such little data, in comparison to the number of free parameters?' -- we see that although the neural network may have many tunable weights, high correlation makes most of them largely redundant.   So, although a neural network may have millions or billions of parameters, they are effectively exponentially smaller.  This low dimensional manifold emerges naturally, and by forcing additional correlation with a shortcut connection, we further increase the effective redundancy and speed training.  By extension, in protein folding or the neural connectome, connecting distal components of the system forces correlation of the intervening amino acids or neurons, respectively.  So, although the space of possible arrangements may be combinatorically large, long-range connections decrease the effective space of possible arrangements exponentially.

\section{Acknowledgements} PS would like to thank E. Yeung and P. Mehta for very helpful discussions and comments. The work of PS was partially supported by the Pacific Northwest National Laboratory Laboratory Directed Research and Development (LDRD) Project ``Multiscale modeling and uncertainty quantification for complex non-linear systems". The work of NOH was supported by PNNL's LDRD Analysis in Motion Initiative and Deep Learning for Scientific Discovery Initiative.

\bibliography{hodas_stinis}
\bibliographystyle{siam}

\end{document}